%% file: arxiv_report.tex
\documentclass[11pt,a4paper]{article}
\usepackage[T1]{fontenc}
\usepackage{lmodern}
\usepackage[a4paper,margin=1in]{geometry}
\usepackage{amsmath,array,booktabs,longtable,pdflscape,tikz}
\usetikzlibrary{arrows.meta,positioning,calc}
\usepackage[unicode,hidelinks]{hyperref}
\hypersetup{pdftitle={From Volpiano to Pixels: 30,000 Lines of OMR Ground Truth for Medieval Chant},pdfauthor={Robert Klugseder}}
\date{}

\title{From Volpiano to Pixels: 30,000 Lines of OMR Ground Truth for Medieval Chant}

\author{\href{https://www.oeaw.ac.at/de/acdh/institut/team/mitarbeitende/robert-klugseder}{Robert Klugseder}\\{\normalsize Austrian Academy of Sciences (ACDH), Vienna}\\{\normalsize\href{mailto:robert.klugseder@oeaw.ac.at}{robert.klugseder@oeaw.ac.at}}}

\begin{document}

\maketitle

\begin{abstract}

Optical music recognition (OMR) for medieval chant lacks training data linking manuscript images to musical symbols across notational families. Using existing Volpiano transcriptions from the Cantus Database \cite{cantusdb}, we created 30,567 image--target pairs from ten sources in eleven exports, including 339 verse fragments. Their target sequences contain 1,952,143 tokens. The pipeline locates staff systems, infers clefs and maps known pitches to staff positions. Six detector families handle the different layouts and notations. Where heuristics failed, two small U-Nets learned staff lines and clef letters from manual annotations. Reviews of sampled export lines found zero to four pitch or clef errors, but used different error definitions and do not provide a uniform audit of the corpus. A CRNN trained from scratch achieved a symbol error rate of 9.6 \% on 3,086 test lines. The earlier model, trained mainly on Austrian sources of the 15th and 16th centuries, scored 45.3 \%. In three separate training runs, each excluding one manuscript, error rates on the excluded manuscript were 21.5--31.0 \%. Model-assisted alignment recovered a further 2,686 lines.

\end{abstract}

\input{figure_pipeline.tex}

\input{figure_encoding.tex}

\section{Problem}

\label{sec:1}

A chant OMR model reads a staff image and outputs musical symbols. Training requires images paired with reference transcriptions, often called ground truth. Creating these labels by hand is costly.

We began with a convolutional recurrent neural network (CRNN), which combines image features with sequence recognition. Jorge Calvo-Zaragoza's team at the University of Alicante's Pattern Recognition and Artificial Intelligence Group (PRAIG) \cite{praig} trained the model on ground truth that the author had prepared manually for Austrian chant manuscripts. During this project, the model was available only in ONNX, an exchange format for trained models.

Our copy of the Cantus Database, dated 16 February 2026, contains Volpiano strings \cite{volpiano} for 57,085 chants with image links. Volpiano encodes pitches as ASCII characters and marks word, syllable, neume and line boundaries (\texttt{7}). We can obtain training labels by aligning these transcriptions with the images, after checking each source for omissions, transcription errors and melodic variants.

The closest existing resource is PRAIG's AMNLT \cite{amnlt}, released on Hugging Face under CC BY-NC 4.0. It contains three real-manuscript corpora: Einsiedeln, Salzinnes and Solesmes, with 1,816, 2,965 and 854 staff excerpts. GregoSynth adds 126,579 synthetic staves rendered by GregorioTeX from real GABC transcriptions \cite{gabc}. We aim to add real manuscript images from ten CANTUS sources, covering notational families that are not yet fully represented in OMR training data.

This work draws on the author's experience with Transkribus Text2Image, where a handwriting text recognition (HTR) model produces a rough reading to align an existing transcription with the page. The author found the tool error-prone and worked with a team to improve the workflow for plain text. For chant, notation, lyrics, clefs and chant boundaries must all agree. The clef should accommodate the chant's full range with as few ledger lines as possible. Model readings give us the first points of correspondence.

\section{Data and target encoding}

\label{sec:2}

Volpiano encodes absolute pitches (\texttt{c} = C3, \texttt{f} = F3, \texttt{k} = C4), liquescence as capitals, section bars and the transcriber's line breaks. Our target tokens instead record staff positions, counted from the bottom line (\texttt{l1 z1 l2 \ldots{} l5}, with no zero position). Here \texttt{l} denotes a line and \texttt{z} a space between lines. Clef tokens include \texttt{c1\ldots{}c5}, \texttt{f1\ldots{}f4} and written multi-letter forms such as \texttt{f2/c4}, \texttt{d2/f3} and \texttt{f2/a3/c4/e5}, alongside separators and bars.

The internal encoding converts a pitch to a half-line counter:

\[q = \operatorname{index}(\text{pitch}) - \operatorname{index}(\text{clef pitch}) + 2(L-1),\]

where L is the clef line counted from the bottom. A documented round-trip test converts all 61,244 non-empty Volpiano strings in the snapshot to this encoding and back, using seven reference clefs. The test confirms that the internal conversion preserves the transcription. In the exported targets, however, separators become \texttt{,}, section bars become \texttt{L}, and liquescence is omitted.

The CANTUS inventory's folio and sequence information gives the order of chants on the pages where they start. In Volpiano, \texttt{7} marks a line break and \texttt{77} continues a chant on the next page. We obtain the images through IIIF and resize them to a working width of 1,600 px. This sometimes enlarges smaller originals.

\section{Pipeline}

\label{sec:3}

Each source follows four steps:

\begin{enumerate}

\item Check how many chants have Volpiano, how many contain \texttt{7}, and how many lines occur per page, including the most frequent line count.

\item Download the images and resize them.

\item Detect the staff systems.

\item Export the target sequences in \texttt{gt.txt} and their provenance and method-specific confidence information in \texttt{meta.tsv}.

\end{enumerate}

Clefs and line boundaries come from model readings, colored rules, marginal letters, lyrics alignment or manual annotations. When aligning from notation-model readings, we infer a clef sequence and refine the change positions with connectionist temporal classification (CTC). CTC scores symbol sequences without needing a location label for every symbol. Section~\ref{sec:5} describes the alignment of transcription lines with image systems.

\section{Detector families}

\label{sec:4}

\begin{itemize}

\item For ruled staves that the notation model can read, a filter makes thin lines more visible. A comb-shaped filter tests their spacing, and a grid fitted across the page locates the systems.

\item The red line is always an F line, but its position within the staff can vary. A text anchor and a staff-spacing model estimate its line number. We accept that estimate only when the two agree.

\item Where staff lines are faint, the text line anchors the system. The pipeline then measures the color of each staff line.

\item In lyrics alignment, colored F/C lines locate the systems and the lyrics determine line boundaries. HTR reads the page text for A-KN 1018 and 1012; this is matched against the database full text. A-KN 1010 has no complete page text, so individual verse texts are matched to portions of the recognized text.

\item Where brown staff lines provide no color cue, F/C/D/E/G letters at the line start supply the pitch reference. These marginal letters can appear singly or in groups.

\item Heuristic variants detected about 52 \% of hand-marked systems. We trained a U-Net, an image segmentation network with about half a million parameters, on 235 hand-drawn systems from 23 pages. It located 118 of 124 systems in held-out material within 0.3 staff spaces of the reference.

A second U-Net of similar size learned clef letters from 31 annotated pages. Template matching had found 15 of 101 reviewed marginal clefs. On a separate review set, the learned detector gave 92 correct and 4 incorrect readings, with 18 missing. These results come from different review sets, so they cannot measure a controlled gain in accuracy.

\end{itemize}

\section{Synchronizing text, melody and image}

\label{sec:5}

Each transcription segment needs an image line and a pitch reference that matches the written clef. Several chants can share one line crop, with their boundaries marked in the target sequence. If no marginal letter is recognized, the export uses a default clef form.

\textbf{Lines.} Using notation-model readings, a dynamic program aligns the page's Volpiano lines with the detected systems in their original order. It finds the mapping with the lowest total cost for the page, allowing four operations: match one line to one system, merge two transcription lines into one system, split one transcription line across two systems, or skip a line or system. With A as the normalized alignment cost defined below, a match costs A, a merge or split costs 1.5\ensuremath{\cdot}A + 0.3, and a skip costs 0.8. A missing or surplus line break can thus be handled without shifting the rest of the page.

Other methods require the same number of transcription lines and systems, or align through the lyrics. Where \texttt{7} is unreliable, HTR reads the lyrics and the pipeline matches them against the database full text. Volpiano's word and syllable separators place the melody at the corresponding boundaries.

\textbf{Chant changes inside a line.} CantusDB inventories give the order of chants on each manuscript page. Reliable \texttt{7} markers in Volpiano or lyrics alignment tell us which chants share a line. When using marginal letters for automatic alignment, we estimate each chant boundary's horizontal position from the cumulative note count.

\textbf{Clef prediction.} The notation-based alignment tests seven clefs: \texttt{c1\ldots{}c4} and \texttt{f2\ldots{}f4}. It favors sequences needing fewer ledger lines, although scribes do not always follow this preference. The calculation uses the following main costs:

\begingroup
\setlength{\tabcolsep}{4pt}
\renewcommand{\arraystretch}{1.16}
\begin{longtable}{@{}>{\raggedright\arraybackslash}p{0.66\linewidth}>{\raggedright\arraybackslash}p{0.30\linewidth}@{}}
\caption{Costs used in clef prediction.}\label{tab:costs}\\
\toprule
\textbf{Event} & \textbf{Cost} \\
\midrule
\endfirsthead
\toprule
\textbf{Event} & \textbf{Cost} \\
\midrule
\endhead
\bottomrule
\endfoot
Note outside z\ensuremath{-}1\ldots{}z4 on a four-line staff, or z\ensuremath{-}1\ldots{}z5 on a five-line staff & 0.1 per half-line step beyond the range \\
Clef change within a neume & Forbidden \\
Clef change at a neume separator & 2.0 \\
Clef change at a syllable or word separator & 1.5 \\
Clef change at a new-chant marker & 0.3 \\
Clef at a forced system boundary & No transition penalty \\
Insert or delete a clef in the alignment & 0.5 \\
Insert or delete a note in the alignment & 1.0 \\
\end{longtable}
\endgroup

The alignment cost includes the model's predicted notes and simple clef tokens. Inserting or deleting a clef costs less than inserting or deleting a note, because the model often misses clefs. We add the edit costs and the penalties for ledger lines and clef changes, then divide by the length of the longer predicted or inferred note-and-clef sequence. The result is A, the normalized alignment cost. It differs from the interval NED used for screening in section~\ref{sec:8}.

The lowest-cost clef sequence gives a hypothesis about pitch placement. The written clef form still needs evidence; marginal letters provide the primary evidence where they are available.

For automatic alignment using letters, we accept a mid-line clef change only at a chant boundary. The boundary must lie within 0.2 line widths of the letters and be clearly closer than the alternatives. Each part must also pass a pitch-range filter, allowing l\ensuremath{-}1 to l5 on a four-line staff. Changes entered by the expert can instead specify a note index within a chant.

\section{The expert in the workflow}

\label{sec:6}

Before we calibrate a detector, the expert describes the page layout in a questionnaire. In a browser editor, the expert draws reference lines and clef marks, then checks the target symbols overlaid on the images. Each pipeline compares the assignment, displayed tokens or prediction version to determine whether an earlier verdict still applies.

The expert uses three interfaces: one to draw staff geometry, one to review pages by clicking on systems, and one to review only new or changed lines. The drawn geometry doubles as training data. Notes such as "clef change to f3 after 21 notes" are parsed and applied directly as corrections to the ground truth.

\begin{landscape}

\section{Sources and results \cite{images}}

\label{sec:7}

\begingroup
\setlength{\tabcolsep}{4pt}
\renewcommand{\arraystretch}{1.16}
\small
\begin{longtable}{@{}>{\raggedright\arraybackslash}p{3.6cm}>{\raggedright\arraybackslash}p{4.4cm}>{\raggedright\arraybackslash}p{3.9cm}>{\raggedleft\arraybackslash}p{1.4cm}>{\raggedright\arraybackslash}p{6.0cm}>{\raggedleft\arraybackslash}p{2.0cm}@{}}
\caption{Sources and reviewed results.}\label{tab:sources}\\
\toprule
\textbf{Source} & \textbf{Provenance} & \textbf{Notational family} & \textbf{Lines} & \textbf{Errors/verdicts (Wilson UB)} & \textbf{Harvested (model)} \\
\midrule
\endfirsthead
\toprule
\textbf{Source} & \textbf{Provenance} & \textbf{Notational family} & \textbf{Lines} & \textbf{Errors/verdicts (Wilson UB)} & \textbf{Harvested (model)} \\
\midrule
\endhead
\bottomrule
\endfoot
A-KN 1010 (12th c.) & Augustinian canonesses, Klosterneuburg & Klosterneuburg notation & 339 & 101 verse verdicts, 16 misassigned &  \\
A-KN 1012 (12th c.) & Augustinian canonesses, Klosterneuburg & Klosterneuburg notation & 1,175 & 4/206 (4.9 \%) & 706 \\
A-KN 1018 (14th c.) & Augustinian canonesses, Klosterneuburg & Klosterneuburg notation & 2,558 & 1/202 (2.8 \%) &  \\
A-Wn 1799** (early 13th c.) & Cistercian abbey of Rein & Cistercian notation & 3,137 & 0/34 definite errors; 1 unclear (conservative 1/35: 14.5 \%) &  \\
CH-E 611 (14th c.) & Benedictine abbey of Einsiedeln & square notation & 7,413 & 2/99 (7.1 \%) &  \\
CH-Fco 2 (late 13th c.) & unknown Franciscan house & square notation & 6,000 & 0/65 (5.6 \%) &  \\
CH-P 18 (c. 1160) & unknown Premonstratensian house & early Messine staff notation & 1,826 & 2/200 (3.6 \%) & 755 \\
D-KA Aug. LX (c. 1200 and 13th/14th c.) & Benedictine abbey of Zwiefalten & German Gothic chant notation & 3,823 & 1/81 (6.7 \%) &  \\
D-KA addendum (15th c.) & Benedictine abbey of Zwiefalten & German Gothic chant notation & 423 & 0/70 pitch (5.2 \%), 4 clef forms &  \\
D-KNd 1161 (13th c.) & unknown Cistercian house & German Gothic chant notation & 1,213 & 3/85 (9.9 \%) &  \\
F-Pn lat. 12044 (early 12th c.) & Benedictine abbey of Saint-Maur-des-Fossés & early French staff notation & 2,660 & 1/210 (2.6 \%) & 1,225 \\
\textbf{Total} &  &  & \textbf{30,567} &  & \textbf{2,686} \\
\end{longtable}
\endgroup

\end{landscape}

"Harvested" lines were recovered with our own model (section~\ref{sec:10}) and are additional to the main line count. Their normalized alignment cost A (section~\ref{sec:5}) is at most 0.12. Detected letters or the red-line layout support the clef form; otherwise it is marked as computed. Samples of 50 lines per source found one pitch error for F-Pn 12044 and none for CH-P 18 or A-KN 1012. The Wilson upper bound is 10.5 \% for the sample with one error and 7.1 \% for each sample without an error.

The reviews count different errors: pitches only for the addendum, pitches and written clef forms for Paris, and verse assignments for A-KN 1010. The Paris and Klosterneuburg reviews also cover only lines admitted by the export rules. All quoted bounds are the upper endpoints of two-sided 95 \% Wilson intervals for the reviewed subsets. They cannot serve as a uniform error bound for the complete corpus.

\section{Evaluation}

\label{sec:8}

For initial screening, we read about eight pages and convert each system's predicted notes into intervals between consecutive pitches. We compare these intervals with the best-matching Volpiano line on the same page and on a comparison page. Normalized edit distance (NED) counts insertions, deletions and substitutions, divided by the length of the longer sequence. The median distances range from 0.089 versus 0.620 in the best case to 0.756 versus 0.771 when the model gives no usable signal.

The Paris screening log includes the comparison page among its nine pages, so one page is compared with itself. This rules out using the screening results as an independent benchmark. The expert checks the exported labels against the images. Even a poor model reading can lead to correct labels when hand-drawn references and clef letters supply the geometry. The sources with the lowest reviewed error counts are also easy to read, which prevents us from isolating the effect of any one method.

AMNLT provides a musical transcription of CH-E 611 that we could compare with our labels. We matched 1,519 systems; the transcriptions differ in 357. Of 40 randomly sampled differences, the expert judged 39 against the images. AMNLT was wrong in 26 cases, our ground truth in 10, and both in 3. Our ground truth was therefore wrong in 13 of the 39 judged differences. Multiplying this proportion by the share of differing systems gives an estimated error rate of 7.8 \% across matched systems, with a scaled Wilson interval of 4.8--11.5 \%. The estimate assumes that agreeing systems are correct. It misses errors shared by both transcriptions and may underestimate the error if difficult lines are absent from the matching.

We separately estimated clef-change errors at 2.9 \%, including one additional case recorded in an expert note without a formal verdict. Other sampled errors were wrong individual pitches in seven lines. The expert also observed that AMNLT omits liquescent notes recorded in Volpiano and often confuses pes and clivis. In the expert's assessment, these differences show that the AMNLT labels were not copied from the CantusDB Volpiano and point to automatic recognition.

\section{Lessons}

\label{sec:9}

\begin{enumerate}

\item Interval NED misses shifts of the entire staff. Moving the staff by one line leaves pitch intervals unchanged, so its geometry needs visual checking before it supports a quality claim.

\item Clef changes can escape review. Because Volpiano encodes no change within a chant, an export with one clef per line can be incomplete. Random reviews, targeted checks and the AMNLT comparison exposed this weakness. The expert marked mid-line changes and their letters on 73 selected pages of F-Pn 12044. In the first round of 23 pages, 6 of 18 systems had unsupported changes inferred from alignment alone. We then required the letter detector to find a letter group of the same pitch. On the remaining 50 pages, changes were correct in 27 of 30 systems. Of 38 measured changes, 27 lay near an estimated chant boundary. About 12 \% of systems with a chant boundary carried a change.

\item Transcription conventions affect yield. Differentiae, repetenda and incipits appear in some sources and are absent in others. Checking them against the images substantially changed the yield, including about 870 initially blocked lines in A-KN 1018.

\item Notation style appears relevant to transfer. In D-KA Aug. LX, the model reads the later addendum much better than the c. 1200 main part (NED 0.089 versus 0.42). Yet 14th-century Klosterneuburg yielded little usable signal despite its proximity to the training material: about 10 to 20 notes read per system against about 35 written. Detector and crop also differed. The comparison suggests a possible effect of notation style, but cannot measure it.

\item Annotated examples help when heuristics stall. Twenty-three annotated pages sufficed for the line model and 31 for the clef model after heuristics stalled near 50 \%.

\end{enumerate}

\section{Training and model-assisted alignment}

\label{sec:10}

The available ONNX model and the released training architecture reduce image width differently, which prevented direct fine-tuning. We therefore trained a CRNN with CTC from scratch on the exported lines. The initial dataset index used for evaluation contains 30,586 lines from an earlier export snapshot than the one in section~\ref{sec:7}. We intended to split each source by folio, but a parsing error placed fragments from one A-KN 1010 folio in both training and test. The converter has since been corrected. The reported evaluations still use the original index.

On the index's 3,086 test lines, the initial model achieved a symbol error rate (SER) of 9.6 \%; the earlier model scored 45.3 \%. SER sums token insertions, deletions and substitutions across all test lines and divides by the total number of reference tokens. The models read the same crops, but have different vocabularies. Since the earlier model cannot output all target tokens, the SER comparison measures compatibility with our exported labels as well as recognition performance. Interval NED excludes clef and separator tokens. It also improved, from 0.868 to 0.034 for F-Pn 12044 and from 0.667 to 0.069 for CH-P 18.

We tested transfer in three separate training runs, leaving out one source entirely in each run. On the excluded source, SER was 21.5 \% for A-KN 1012, 31.0 \% for D-KNd 1161 and 24.8 \% for F-Pn 12044, compared with 56--81 \% for the earlier model. These three cases suggest that the model can help with additional manuscripts, but do not establish an expected error rate for all new sources.

We used the initial model to recheck lines that the pipeline had withheld. To accept a line, we require a normalized alignment cost A (section~\ref{sec:5}) of at most 0.12. The alignment infers the final clef sequence, with the model's predicted simple clef tokens contributing to the cost. The final sequence is not copied directly from the model output. Detected letters or the red-line layout establish the written clef form where available; otherwise the form is marked as computed. We recovered 2,686 lines in three sources. Samples of 50 lines per source found one pitch error in total (section~\ref{sec:7}).

We then retrained the model on the extended corpus, assigning each folio to one split only. We excluded harvested lines from validation or test folios, along with two A-KN 1010 fragments whose folio is in the test set. On the same 3,084 test lines, the retrained model reaches an SER of 9.56 \%, equal to the initial model. Its reading of initial clef groups improved markedly: the share of lines with the correct initial clef token rose from 43 \% to 90 \% for A-KN 1012, from 48 \% to 64 \% for F-Pn 12044 and from 19 \% to 33 \% for D-KNd 1161. Single-letter clefs became less accurate. For CH-P 18, the share fell from 25 \% to 9 \%. This metric checks only the initial token, including a multi-letter form where present, and does not assess changes within a line.

The ground truth favors lines that the pipeline could resolve, leaving faint pages and rare clef forms under-represented. Error definitions vary by source, and several rules were developed on the same material used for review. Each target gives an ordered sequence of staff positions within a line crop, without coordinates for individual noteheads. Missing clef changes are not yet detected automatically.

\section{Roles of the expert and AI assistant}

\label{sec:11}

The domain expert worked with an AI coding assistant (Claude Fable 5.1 and Opus 5.5 in Claude Code) on the pipeline, review tools and this report. The repository contains about 300 commits, 29 Python files for the pipeline and additional review utilities.

The assistant worked largely autonomously: it screened sources, implemented detectors, ran the full-book passes and built browser review interfaces. It also measured yield and error rates and trained the U-Nets and CRNN. When the heuristics stalled, it proposed learning from the expert's 235 hand-drawn systems. The expert supplied chant research expertise, paleographic readings and annotations, and decided on thresholds, rules and export readiness. Measurements supported the implementation. The expert interpreted the results and accepted the labels.

\input{references.tex}
\end{document}

%% file: figure_pipeline.tex
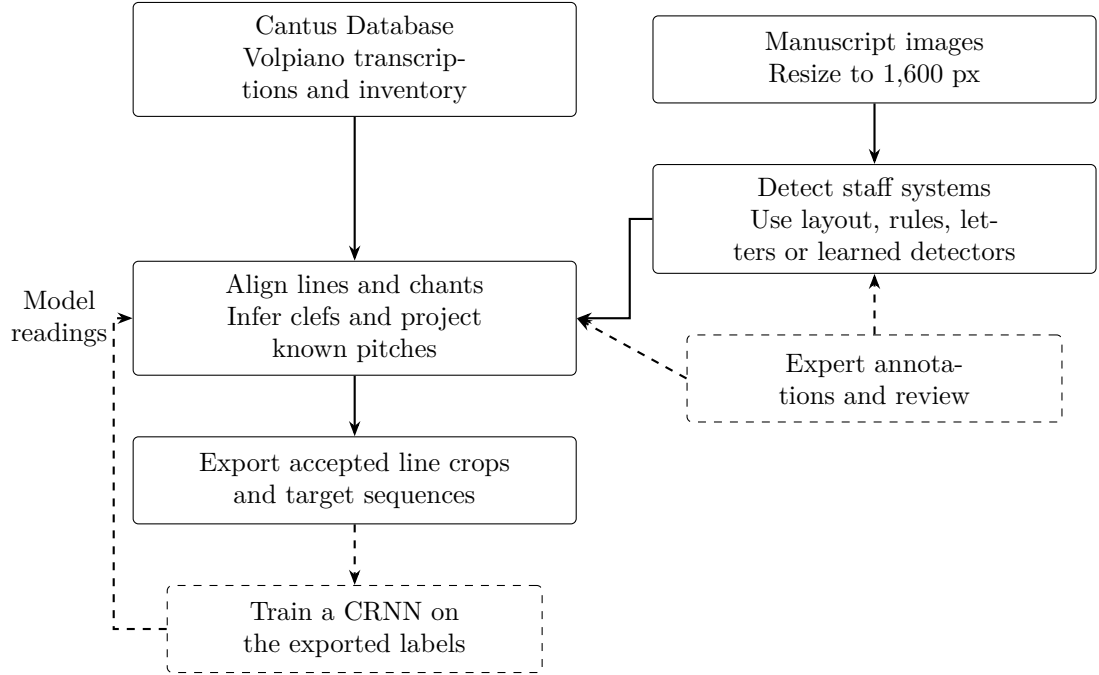
\begin{figure}[tbp]
\centering
\begin{tikzpicture}[
  box/.style={draw, rounded corners=2pt, align=center, text width=5.5cm,
    minimum height=1.15cm, inner sep=5pt, font=\small},
  support/.style={box, dashed, text width=4.6cm},
  arrow/.style={-{Stealth[length=2mm]}, thick},
  node distance=8mm and 10mm
]
\node[box] (transcriptions) {Cantus Database\\Volpiano transcriptions and inventory};
\node[box, right=of transcriptions] (images) {Manuscript images\\Resize to 1,600 px};
\node[box, below=of images] (systems) {Detect staff systems\\Use layout, rules, letters or learned detectors};
\node[box, below=19mm of transcriptions] (alignment) {Align lines and chants\\Infer clefs and project known pitches};
\node[support, below=of systems] (expert) {Expert annotations and review};
\node[box, below=of alignment] (export) {Export accepted line crops\\and target sequences};
\node[support, below=of export] (model) {Train a CRNN on the exported labels};
\draw[arrow] (images) -- (systems);
\draw[arrow] (transcriptions) -- (alignment);
\draw[arrow] (systems.west) -- ++(-3mm,0) |- (alignment.east);
\draw[arrow] (alignment) -- (export);
\draw[arrow, dashed] (expert.west) -- (alignment.east);
\draw[arrow, dashed] (expert.north) -- (systems.south);
\draw[arrow, dashed] (export) -- (model);
\draw[arrow, dashed] (model.west) -- ++(-7mm,0) |- (alignment.west)
  node[pos=0.62, left, align=center, font=\small] {Model\\readings};
\end{tikzpicture}
\caption{Schematic of the workflow described in sections~\ref{sec:3}--\ref{sec:6}
and \ref{sec:10}. Dashed arrows show expert input and model feedback.}
\label{fig:pipeline}
\end{figure}

%% file: figure_encoding.tex
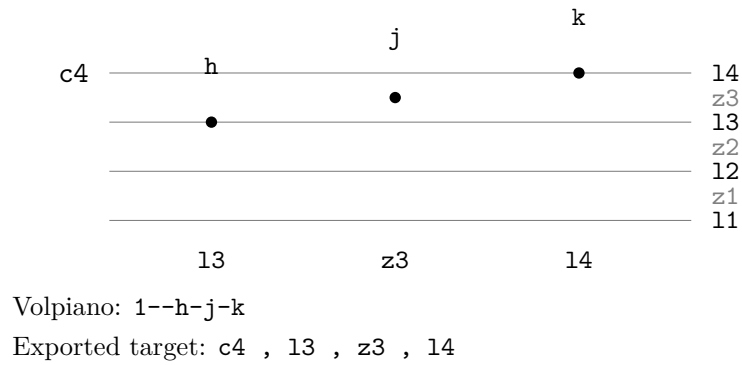
\begin{figure}[tbp]
\centering
\begin{tikzpicture}[x=1.35cm,y=0.65cm, font=\small]
  \node[anchor=west, font=\normalsize\bfseries] at (-1.05,5.1)
    {Known pitches mapped to staff positions};
  \foreach \y/\label in {0/l1,1/l2,2/l3,3/l4} {
    \draw[gray] (0,\y) -- (5.7,\y);
    \node[anchor=west] at (5.8,\y) {\texttt{\label}};
  }
  \foreach \y/\label in {0.5/z1,1.5/z2,2.5/z3} {
    \node[anchor=west, text=gray] at (5.8,\y) {\texttt{\label}};
  }
  \node[anchor=east, font=\normalsize] at (-0.1,3) {\texttt{c4}};
  \foreach \x/\y/\pitch/\token in {1/2/h/l3,2.8/2.5/j/z3,4.6/3/k/l4} {
    \fill (\x,\y) ellipse [x radius=0.075cm,y radius=0.075cm];
    \node[above=5mm] at (\x,\y) {\texttt{\pitch}};
    \node at (\x,-0.8) {\texttt{\token}};
  }
  \node[anchor=west] at (-1.05,-1.8)
    {Volpiano: \texttt{1---h-j-k}};
  \node[anchor=west] at (-1.05,-2.6)
    {Exported target: \texttt{c4 , l3 , z3 , l4}};
\end{tikzpicture}
\caption{Illustrative four-line example of the encoding in section~\ref{sec:2}.
The C clef on line~4 fixes the reference pitch. The tokens \texttt{l} (line) and
\texttt{z} (Zwischenraum, German for space) denote lines and spaces counted from the bottom.}
\label{fig:encoding}
\end{figure}